\def\year{2022}\relax
\documentclass[letterpaper]{article} 
\usepackage{aaai22}  
\usepackage{times}  
\usepackage{helvet}  
\usepackage{courier}  
\usepackage[hyphens]{url}  
\usepackage{graphicx} 
\usepackage{natbib}  
\usepackage{caption} 
\usepackage{booktabs}
\usepackage{makecell}
\usepackage{amsfonts}
\usepackage{multirow}
\usepackage{algorithm}
\usepackage{algorithmic}

\usepackage{newfloat}
\usepackage{listings}
\DeclareCaptionStyle{ruled}{labelfont=normalfont,labelsep=colon,strut=off} 
\floatstyle{ruled}
\newfloat{listing}{tb}{lst}{}
\floatname{listing}{Listing}
\begin{document}
%
%


\title{SDFE-LV: A Large-Scale, Multi-Source, and Unconstrained Database for Spotting Dynamic Facial Expressions in Long Videos}
\author{
    Xiaolin Xu,\textsuperscript{\rm 1}
    Yuan Zong,\textsuperscript{\rm 1}
\thanks{Corresponding authors}
    Wenming Zheng,\textsuperscript{\rm 2 ${\ast}$}
    Yang Li,\textsuperscript{\rm 1} \\
    Chuangao Tang,\textsuperscript{\rm 3} 
    Xingxun Jiang,\textsuperscript{\rm 1}
    Haolin Jiang \textsuperscript{\rm 1}
}
\affiliations{
    $^1$School of Biological Science and Medical Engineering, Southeast University, Nanjing, China \\
    $^2$Key Laboratory of Child Development and Learning Science, Ministry of Education, Southeast University, Nanjing, China \\
    $^3$School of Information Science and Engineering, Southeast University, Nanjing, China \\
    \{xuxiaolin, xhzongyuan, wenming\_zheng, li-yang, tcg2016, jiangxingxun, hljiang\}@seu.edu.cn
}




\maketitle


\begin{abstract}
In this paper, we present a large-scale, multi-source, and unconstrained database called SDFE-LV for spotting the onset and offset frames of a complete dynamic facial expression from long videos, which is known as the topic of dynamic facial expression spotting (DFES) and a vital prior step for lots of facial expression analysis tasks. Specifically, SDFE-LV consists of 1,191 long videos, each of which contains one or more complete dynamic facial expressions. Moreover, each complete dynamic facial expression in its corresponding long video was independently labeled for five times by 10 well-trained annotators. To the best of our knowledge, SDFE-LV is the first unconstrained large-scale database for the DFES task whose long videos are collected from multiple real-world/closely real-world media sources, e.g., TV interviews, documentaries, movies, and we-media short videos. Therefore, DFES tasks on SDFE-LV database will encounter numerous difficulties in practice such as head posture changes, occlusions, and illumination. We also provided a comprehensive benchmark evaluation from different angles by using lots of recent state-of-the-art deep spotting methods and hence researchers interested in DFES can quickly and easily get started. Finally, with the deep discussions on the experimental evaluation results, we attempt to point out several meaningful directions to deal with DFES tasks and hope that DFES can be better advanced in the future. In addition, SDFE-LV will be freely released for academic use only as soon as possible once the paper is accepted.
\end{abstract}

\begin{table*}[t!]
\centering
\renewcommand{\arraystretch}{1.3}
\caption{Comparison between our SDFE-LV and Existing Publicly Available Facial Expression Databases for DFES, where LV and DFE are Abbreviations of Long Videos and Dynamic Facial Expressions.}  
\label{summary}
\begin{tabular}{l|c|c|c|c|c|c}
\hline 
Database & \#LVs & \#DFEs & \makecell[c]{Average Time \\of LV (s)} & Sources  & DFE Type &~~~ Year~~~ \\
\hline 
SMIC-E-VIS & 71   & 71  & 5.9   & Videos Recored in Lab                                & Micro          & 2017 \\
SMIC-E-NIR & 71   & 71  & 5.9 & Videos Recored in Lab                                & Micro     &     2017 \\
SMIC-E-HS  & 157   & 161  & 5.9   & Videos Recored in Lab                                & Micro          & 2017\\
MEVIEW     & 40    & 40   &  3   & Poker Game Videos, Interviews                               & Micro    &  2017    \\
CAS(ME)$^2$ & 97   & 357  & 87   & Videos Recored in Lab                                & Micro\&Macro          & 2018\\
SAMM-Long     & 147  & 502  & 16 & Videos Recored in Lab                                & Micro\&Macro           & 2019 \\ 
SMIC-E-Long & 162   & 166  & 22   & Videos Recored in Lab                                & Micro\&Macro &2021 \\
CAS(ME)$^3$    & 1300  & 4086  & 98.17 & Videos Recored in Lab                                & Micro\&Macro          & 2022 \\ \hline\hline
\textbf{SDFE-LV}       & 1191 & 2420 & 57.94  & \makecell[c]{Interviews, Reality Shows, Speeches,\\Operas, Movies, Doumentaries,\\ We-Media Short Videos, Others}  & Micro\&Macro & 2022 \\
\hline 
\end{tabular}
\end{table*}

\section{Introduction}

The research of dynamic facial expression spotting (DFES) aims to enable the computers to automatically extract complete facial expression sequences from a long video containing one or more facial expressions by spotting their oneset and offset frames~\cite{shreve2009towards}. It is an essential prior step for lots of high-level facial expression analysis tasks, e.g., dynamic facial expression recognition (FER), micro-expression recognition (MER), and facial action unit detection. Hence, DFES determines whether the above analysis techniques can be usable in practice. The early research of DFES may be tracted to the work of~\cite{de2007temporal}, in which De la Torre et al. proposed a two-steps approach including shape and apperance feature clustering and temporal cluster grouping to extract the temporal segments of generalized facial gestures instead of only dynamic faical expressions from spontaneous facial behavior long videos. Sherve et al.~\cite{shreve2009towards,shreve2011macro,shreve2014automatic} investigated automatic spotting of facial expressions including both macro- and micro-expressions by presenting a series of strain pattern based methods. 

However, it is noted that the datasets collected in the above pioneer works were not subsequently released and hence lack of benchmark databases hindered the progress of DFES research in the first few years after that. It can be seen from existing dynamic facial expression databases, e.g., CK+~\cite{intro:ck+}, CAER~\cite{intro:caer}, DFEW~\cite{intro:dfew}, and FERV39k~\cite{intro:ferv39k} that most of them all support the evaluation of recognition or action unit detection methods. In other words, in these databases the DFES step was manually performed in advance and released facial expression video clips are all the ones starting from the onset and ending at the offset. Fortunately, researchers from MER have came to realize the importance of DFES in recent years and built several databases to support the research of DFES. 

Until 2017, the first publicly available DFES database called SMIC-E~\cite{li2017towards} consisting of three subsets, i.e., SMIC-E-HS, SMIC-E-VIS, and SMIC-E-NIR, was released. SMIC-E is an extension of SMIC~\cite{pfister2011recognising}, 
which is also the first publicly available spontaneous micro-expression database for recognition tasks. Since then, lots of databases are one by one released including MEVIEW~\cite{intro:meview}, CAS(ME)$^2$~\cite{intro:casme2}, SAMM-Long~\cite{intro:samm-lv}, SMIC-E-Long~\cite{intro:smic-e-long}, and CAS(ME)$^3$~\cite{li2022cas}. We summarize the statistical information of these facial expression databases for DFES in Table~\ref{summary}. From Table~\ref{summary}, it is clear to see that although existing databases paid more attention to spotting micro-expressions (usually lasting within 0.5s) from long videos, macro-expressions (alternative name of ordinary dynamic facial expressions usually lasting over 0.5s) were also considered in most of them due to the symbiosis between macro- and micro-expressions in real life~\cite{ekman1992facial}. Note that compared with the ordinary dynamic facial expressions, spotting micro-expressions is a more tough task due to their two important characteristics, i.e., low intensity and short duration. 

It should be also pointed out that almost all of existing micro-expression databases supporting DFES were collected in the lab-controlled environment. This thus leads to that ordinary DFES (or say macro-expression spotting) methods developed based on these databases may not meet the requirement of applications in practical unconstrained environment because of the gap between lab-controlled and unconstrained ones. Unlike the lab-controlled enviroment, the task of DFES as well as its downstream tasks, e.g., FER, in the uncontrained one has lots of challenges such as head posture changes, occlusions, illuminations, and temporal scale variations of face, resulting the sharp increase of the task's difficulty. As Table~\ref{summary} shows, only the long videos of MEVIEW~\cite{intro:meview} were collected from unconstrained media sources including Poker game and interview videos. Despite of this, the average duration of long videos in MEVIEW only reaches 3 seconds, which is far from the real-world scenes. Moreover, MEVIEW~\cite{intro:meview} merely consists of 40 long videos, whose sample number is at a unsatisfactory level. Over past several years, researchers of FER and other facial behavior analysis have already shifted their focus from the lab-controlled enviroment to the unconstrained one and presented lots of large-scale database of facial expressions in the wild to further advance the research of FER and facial behavior analysis. Consequently, it is also urgently needed to build a large-scale unconstrained database to support the research of DFES such that furture DFES techniques can better fit FER ones in real-world scenes.

To fill the gap, in this paper we present a large-scale, multi-source, and unconstrained database called SDFE-LV for spotting dynamic facial expressions in long videos. Inspired by most of existing unconstrained facial expression databases for recognition tasks, we first recruited 50 undergraduates and postgraduates familiar with video editing softwares to collect 
long videos containing one or more complete facial expressions from multiple real-world/closely real-world media sources including interviews, reality shows, speeches, operas, movies, doumentaries, we-media short videos, and others. These long videos cover lots of challenging situations in real life, e.g., the above mentioned head posture changes, occlusions, illuminations, and temporal scale variations of face. Then, crowdsourcing was used to label both the onset and offset frame indices of each dynamic facial expression in its corresponding long video for independent five times to determine the final onset and offset grountruth information. Finally, extensive benchmark evaluation experiments by using lots of recent-state-of-the-art deep spotting methods were carried out and discussed. We also summarize the statistical information of SDFE-LV in Table~\ref{summary} such that the readers can better understand its characteristics from the comparison between our SDFE-LV and existing spotting databases. In addition, it is also worthy to mention that to the best of our knowledge, SDFE-LV is the first large-scale unconstrained facial expresson database desinged for spotting tasks. It will be released as soon as possible once this paper is accepted. We sincerely hope SDFE-LV database can truly advance the research of DFES in the future.

\section{Database}
\begin{figure*}[t]
\centering
\includegraphics[width=0.98\linewidth]{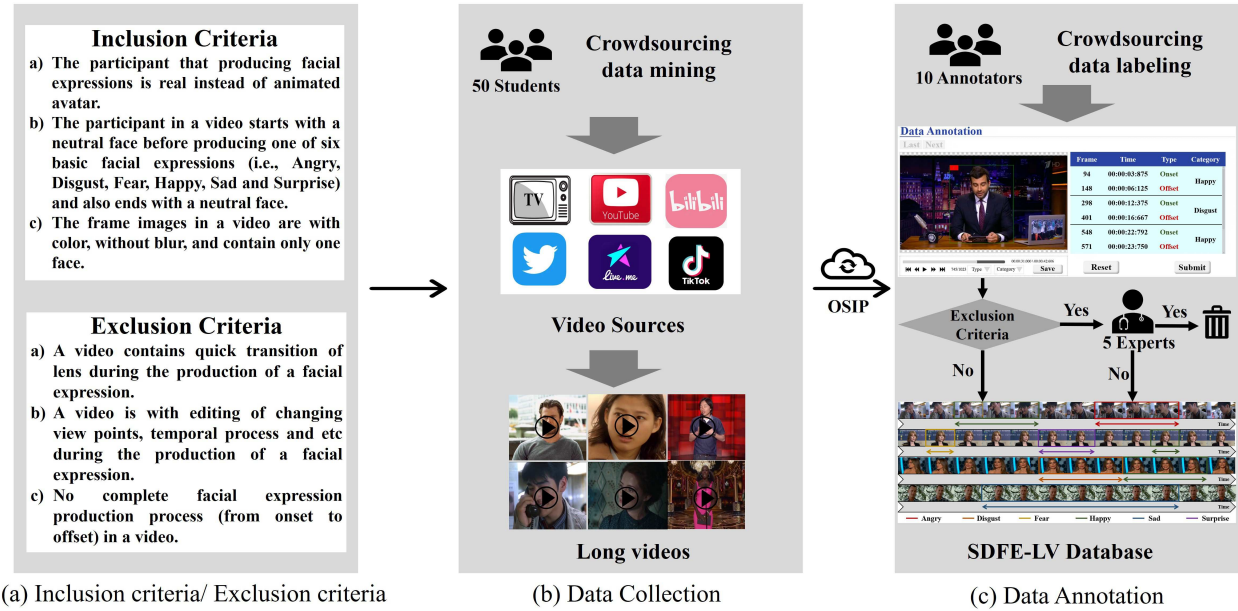}
\caption{The construction process of SDFE-LV database.}
\label{fig:flowchart}
\end{figure*}

This section details the construction process of our SDFE-LV database, including inclusion and exclusion criteria for mining videos from the Web, crowdsourcing-based data collection and data annotation, which is shown in Figure \ref{fig:flowchart}.

\textbf{Inclusion and Exclusion Criteria}: Before data collection, we enacted the inclusion and exclusion criteria for mining videos from the Web.

The \textbf{inclusion criteria} are as follows:

\textbf{a.} The participant that producing facial expressions in a video is real instead of animated avatar;

\textbf{b.} The participant in a video starts with a neutral face before producing one of six basic facial expressions (i.e., angry, disgust, fear, happy, sad and surprise) and also ends with a neutral face;

\textbf{c.} The frame images in a video are with color, without blur, and contain only one face.

The following is \textbf{exclusion criteria}:

\textbf{a.} A video contains quick transition of lens during the production of a facial expression;

\textbf{b.} A video is with editing of changing view points, temporal process and etc. during the production of a facial expression;

\textbf{c.} No complete facial expression production process (from onset to offset) in a video;


\textbf{Data Collection}: Following the inclusion and exclusion criteria, we employed 50 students at university to mine the facial expression videos. During the data collection, the following issues were taken into consideration.

\textit{Multi-Source}: The videos of interviews, reality shows, speeches, operas, movies, documentaries, we-media short videos, and others from several
real-world/closely real-world media sources, including Youtube, Bilibili, Live-me, Tiktok and etc., were included in this database. In comparison with AFEW \cite{intro:afew} and CAER \cite{intro:caer}, in which only single source (e.g., movies or TV shows) was included, our SDFE-LV furthers the progress of dynamic facial expressions in the real world.

\textit{Avoiding duplication}: The collection of facial expression samples from the Internet is different from those databases collected in the lab, because the participants were controllable in the latter environment, e.g., the subject identities and the number of subjects were known. However, collecting large-scale databases from the Web may appear high duplication of samples if this issue is not regarded. As a result, it is not convincible to develop a robust facial expression analysis model based on a database with high duplication of samples. An online shared interactive platform (OSIP) was developed to manage the collected videos. A crowdsourcing student can find the detailed information, i.e., name of samples, storage address, source link, data type and etc., of previously collected samples on the OSIP. In this way, the duplicated samples will not be taken into consideration.

\textit{Sample diversity}: The long-tail phenomenon has been proved in most databases, including object classification databases and facial expression classification databases. For example, the happy samples appeared with a higher rate while disgust and fear samples appeared with the lowest rates in 6 basic facial expressions. In addition to the diversity issue in specific facial expression classes, long videos containing over 5 complete facial expression producing processes are scarce. To mitigate these issues, we provided crowdsourcing students with extra rewards if they find a sample in the fear or disgust category during data collection. So is multiple facial expression producing processes in a long video.

After data collection, we finally obtained 1,563 long videos by 50 crowdsourcing students. These videos will be further manually annotated for onset, offset and facial expression categories.

\subsubsection{Data Annotation}
To provide reliable data annotation for DFES, we followed the data annnotation strategy in \cite{intro:dfew} and employed 10 annotators in JD Crowdsourcing \footnote{https://weigong.jd.com/} to mark onset, offset of complete facial expression producing at frame level and rate the complete facial expression producing video clips to be one of six basic facial expression categories. Additionally, five facial expression annotation experts with psychology background and expertise in facial expression understanding were employed to train these annotators and selected 2.5\% of the data in the database for practice.

Before annotation, a marking software shown in Figure \ref{fig:flowchart}(c) was developed to help annotators conduct the annotation task. From the software interface, the annotators can easily forward/backward the video stream to capture the onset and offset and mark corresponding results by the 'Type' menu on the interface. One of six basic facial expressions was then rated for the video segment with the marked onset and offset through the 'Category' menu. The marked results are also shown on the right part of the software interface. 

During data annotation, the OSIP was also visible for the crowdsourcing annotators in JD crowdsourcing. If a long video has been marked with the onset, offset and facial expression category, the corresponding label information containing these three elements will be added for the video on the OSIP. In addition, the inclusion and exclusion criteria were also checked by these annotators. If they find that a sample video does meet the exclusion criteria, the videos will be further checked by aforementioned 5 experts. The expert-level check outcomes a decision of marking facial expression with onset, offset and facial expression categories, or discarding such samples when the samples were regarded as meeting the exclusion criteria.

To improve data annotation efficiency, each video in this database was marked by random 5 in 10 annotators. Finally, we received 1,191 qualified long videos certified by these annotators and 5 experts.

\begin{figure}[htbp]
\centering
\includegraphics[width=\linewidth]{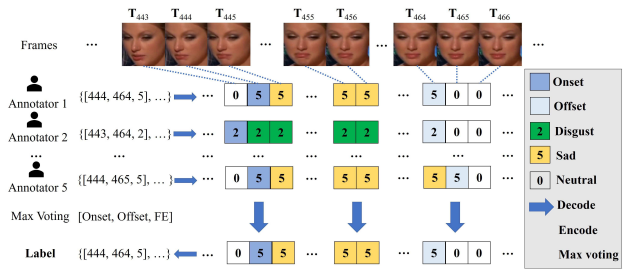}
\caption{The max-voting strategy-based decision for onset, offset and facial expression categories.}
\label{fig:merge}
\end{figure}

For each annotation with the three marked elements, aka, onset, offset and facial expression category, was collected and decoded to frame-level annotation, which is shown in Figure \ref{fig:merge}. Three-element annotations in a long video can be defined as: $\left[[ S^0, E^0, C^0], [ S^1, E^1, C^1] \cdots, [ S^n, E^n, C^n]\right]$, where $S^i, E^i, C^i$ represent the onset, offset, and the category of the $i$-th complete facial expression clip in a long video, and $C \in \{1,2,3,4,5,6\}$, from 1 to 6 are anger, disgust, fear, happy, sad and surprise, respectively, $n$ is the number of complete facial expression producing clips in a video marked by annotators. The final label for a long video was attained by using max-voting strategy for the frame-level decoded annotations from 5 annotators and were further stored in a three-element way.


\subsection{Database Statistics}

\begin{figure}[htbp]
\centering
\includegraphics[width=\linewidth]{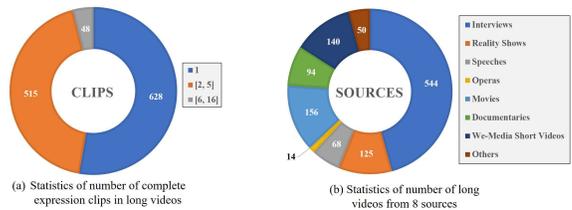}
\caption{Statistic of our SDFE-LV database.}
\label{fig:statistic}
\end{figure}

SDFE-LV is a large-scale, multi-source, and unconstrained facial expression spotting database that contains 1,191 videos with an average duration of 57.94 seconds.
Moreover, each video contains at least one complete expression clip, which can be classified as one of the 6 basic expressions, and the statistical results show that the database contains a total of 2420 complete expression clips.
Influenced by the additional rewards, our SDFE-LV database exhibits extremely strong sample diversity. In Figure \ref{fig:statistic}(a), we show the distribution of the number of complete expression clips in long videos. 
We find that a long video in our database contains at most 16 complete expression clips, which undoubtedly exacerbates the difficulty of expression spotting in our database.
Considering that the video samples in our database come from different sources including interviews, reality shows, speeches, operas, movies, doumentaries, we-media short videos, and others, we count the source distribution of long videos in the database in Figure \ref{fig:statistic}(b). 
We find that long videos from interviews accounted for the majority of the samples in the database with its stable facial images for long periods of time.
\begin{table}[htbp]
\centering
\caption{The distribution of six basic complete expression clips at different durations.}  
\label{distribution}
\begin{tabular}{lcccc} 
\hline
                  & \textbf{[0,0.5]} & \textbf{(0.5,4]} & \textbf{(4,$\infty$)} & \textbf{Total}  \\ 
\hline
Angry    & 3                & 35               & 44             & 82     \\
Disgust  & 3                & 14               & 1              & 18     \\
Fear     & 0                & 2                & 0              & 2      \\
Happy   & 28               & 1200             & 957            & 2185   \\
Sad      & 0                & 21               & 43             & 64     \\
Surprise & 10               & 45               & 14             & 69     \\ 
\hline
Total    & 44      & 1317    & 1059  & 2420   \\
\hline
\end{tabular}
\end{table}
Table \ref{distribution} shows the distribution of six basic complete expression clips under different durations. 
We find that most of the expression clips are concentrated between 0.5 and 4 seconds, but most expressions with anger and sadness last longer than 4 seconds.
In addition, even though we use various incentive measures to motivate crowdsourcing students to search for relatively rare expression clips, there is still a serious long-tail phenomenon in the expression categories in the dataset, which we believe is inevitable during the collection of expression samples, especially for the rare long videos.
In addition, we also made statistics on the distribution of gender and race of all long videos in SDFE-LV \cite{intro:raf-db}.
In summary, there are 57\% male and 43\% female. 
We divided the race groups into 3 broad categories, with Caucasians accounting for 49\%, Asians for 40\%, and Africans accounting for 11\%.

\section{Experiment}
\subsection{Experiment Setup}
\subsubsection{Protocol}
We employed the subject-independent-based training-validation-test protocol to conduct experiment and evaluated the models performance on the test set. The ratio of samples in the training, valiadation and test sets was set to 6:2:2. The validation set was used to guide model training and the model achieved the best performance on the validation set was used to predict the samples in the test set.

\subsubsection{Pre-processing}
The frame images were sliced with using functions in OpenCV toolbox, then the face++ API \footnote{https://www.faceplusplus.com} was utilized to locate facial landmarks and estimate head postures. Subsequently, the facial region in each frame image was cropped with referring to facial landmarks, the cropped facial region of interest (roi) was further resized to the resolution of 256$\times$256 pixels. For large head pose faces where the landmarks were with obvious errors, the landmarks in the past frame and the next frame were used to compute average results for estimating current facial landmarks positions.

\begin{figure}[htbp]
\centering
\includegraphics[width=0.9\linewidth]{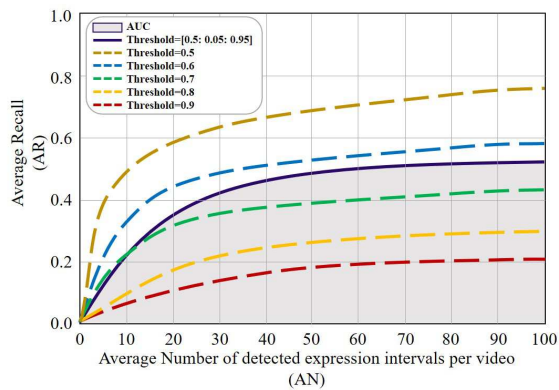}
\caption{Examples of evaluation results on the SDFE-LV database. Dashed lines represent recall when the IoU between $Interval_{DT}$ and $Interval_{GT}$ is over a set threshold, and the solid line is the Average Recall across thresholds [0.5: 0.05: 0.95]. The gray shadow represents the Area Under the Average Recall Curve (AUC).}
\label{fig:auc}
\end{figure}


\begin{table*}[htbp]
\centering
\caption{Comparision results of R-C3D, TURN, BSN, and BMN on the SDFE-LV database in terms of AR@AN and AUC.}  
\label{benchmark}
\resizebox{0.8\linewidth}{!}{
\begin{tabular}{clccccccc}
\toprule 
\textbf{Networks} & \multicolumn{1}{c}{\textbf{Feature Extractor}}           & \textbf{Fashion} & \textbf{AR@1(\%)}   & \textbf{AR@5(\%)}   & \textbf{AR@10(\%)}  & \textbf{AR@50(\%)} & \textbf{AR@100(\%)} & \textbf{AUC}     \\
\midrule 
R-C3D             & C3D                 & Top-down         & 1.09            & 3.71            & 8.07            & 21.44          & 22.30           & 18.29            \\
TURN              & C3D                 & Top-down         & 1.05            & 3.56            & 6.20            & 25.45          & 33.63           & 22.94            \\
BSN               & C3D                 & Bottom-up        & 6.97            & 20.62           & 26.97           & 48.75          & 58.97           & 45.09            \\
BMN               & C3D                 & Bottom-up        & 12.23           & 27.97           & 35.46           & 52.88          & 59.73           & 49.44            \\
\midrule
\midrule  
BMN               & R3D                 & Bottom-up        & 8.00            & 20.72           & 29.15           & 49.09          & 57.77           & 45.63            \\
BMN               & R2Plus1D            & Bottom-up        & 8.95            & 21.86           & 29.76           & 50.33          & 59.00           & 46.89            \\
BMN               & Resnet18-LSTM       & Bottom-up        & 15.50           & 35.10           & 43.15           & 59.54          & 66.92           & 56.43            \\
BMN               & Resnet50-LSTM       & Bottom-up        & 18.28           & 37.82           & 46.31           & 62.07          & 68.67           & 59.03            \\
BMN               & Resnet101-LSTM      & Bottom-up        & 18.36           & 37.80           & 46.36           & 62.52          & 68.75           & 59.19            \\
BMN               & VGG13-LSTM          & Bottom-up        & 17.87           & 36.59           & 46.59           & 61.99          & 67.88           & 58.79            \\
BMN               & VGG16-LSTM          & Bottom-up        & \textbf{19.62 } & 38.00           & 47.73           & 62.46		  & 68.27           & 59.47            \\
BMN               & VGG19-LSTM          & Bottom-up        & 18.75           & \textbf{39.22 } & \textbf{47.90 } & \textbf{62.70} & \textbf{69.41}  & \textbf{59.56 }  \\
BMN               & Two
  Resnet50-LSTM & Bottom-up        & 17.93           & 37.64           & 46.62           & 61.92          & 68.60           & 59.15            \\
BMN               & Two
  VGG19-LSTM    & Bottom-up        & 17.40           & 35.95           & 44.23           & 60.17          & 66.32           & 57.76            \\
BMN               & Two
  C3D           & Bottom-up        & 11.86           & 29.31           & 36.64           & 54.33          & 60.20           & 50.69            \\
BMN               & Two
  TSN           & Bottom-up        & 15.86           & 34.33           & 43.65           & 59.49          & 65.31           & 56.24           \\
\bottomrule 
\end{tabular}}
\end{table*}

\subsubsection{Evaluation Metric}
To ensure that the proposed algorithm can reasonably measure its performance on the SDFE-LV database, we investigate metrics for Target Events Localization works in different fields.
In the micro-expression field, F1-Score\cite{exper:2019megc} was currently regarded as the most widely used MES evaluation metric, which takes both recall and precision into account, expecting to find more Ground Truth expression Intervals ($Interval_{GT}$) with fewer Detected expression Intervals ($Interval_{DT}$). 
However, this evaluation system has no requirements or restrictions on the number of the proposed $Interval_{DT}$, which makes it difficult for each algorithm to make fair comparison under the same number of $Interval_{DT}$ baseline.
Temporal Action Proposals (TAP) task \cite{activitynet}in the action field mainly uses Area Under the Curve (AUC) of the Average Recall $versus$ Average Number of proposals per video (AR@AN) as the evaluation metric, which aims at generating high-quality $Interval_{DT}$ to cover action instances with high recall and high temporal overlap. 
Compared to the F1-Score used by MES, AUC limits the generated number of $Interval_{DT}$ so that different algorithms can be compared more fairly on the same number of $Interval_{DT}$ baseline. 
In addition,  AUC requires $Interval_{DT}$ and $Interval_{GT}$ to have a higher overlap, which is also more practical to real-world scenarios.

Therefore, we choose AUC with more impartiality and practicability as our evaluation metric, which can be calculated as follows:
Firstly, an $Interval_{DT}$ is successfully recalled if it meets the following relationship:
$$
\frac{Interval_{GT} \cap Interval_{DT}}{Interval_{GT} \cup Interval_{DT}} \geq threshold,
$$
where threshold represents the minimum Intersection of Union (IoU) between $Interval_{GT}$ and $Interval_{DT}$ that can be identified as recalled intervals.
Then, we take the number of recalled intervals under thresholds [0.5: 0.05: 0.95] divided by the total number of $Interval_{GT}$  as AR.
At last, we calculate the AR under different Average Numbers of detected expression intervals per video (AN) as AR@AN, where AN varies from 0 to 100, and calculate the Area Under AR@AN Curve as AUC. (shown in Figure \ref{fig:auc})

\subsection{Benchmark Evaluation}

\begin{figure}[htbp]
\centering
\includegraphics[width=\linewidth]{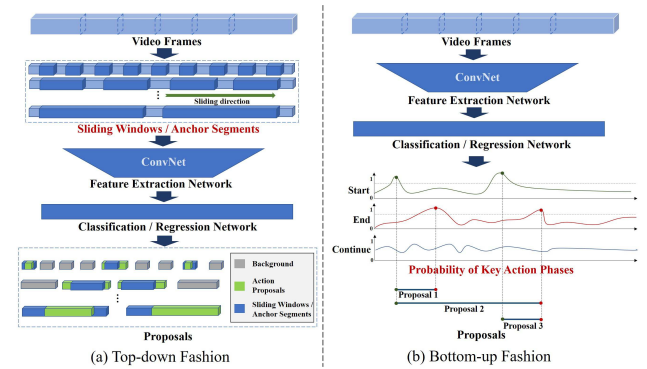}
\caption{Two classic fashions in TAP task.}
\label{fig:fashion}
\end{figure}

In this subsection, recent state-of-the-art deep spotting methods are evaluated as the benchmark for our database.
Considering that algorithms in the field of action have made great strides driven by many researchers, we select the well-performing algorithms in Temporal Action Proposals (TAP) task.
By convention, depending on the fashion of algorithm implementation, algorithms in TAP tasks can generally be divided into top-down and bottom-up fashions. (shown in Figure \ref{fig:fashion})
\textbf{Top-down:} The essence of the top-down algorithm is to pre-define a series of proposals that are regularly distributed (e.g., sliding windows and anchor segments) and then evaluate whether these proposals contain action instances and fine-tune the boundaries slightly. Classic top-down algorithms include TURN\cite{exper:turn}, R-C3D\cite{exper:rc3d}, and TAL-Net\cite{exper:talnet}.
\textbf{Bottom-up:} Bottom-up algorithms first evaluate the frame-level probability of key action phases locally (e.g., start, continue and end phases) and then combine these phases with high probability into action proposals. Classic bottom-up algorithms include SSN\cite{exper:ssn}, BSN\cite{exper:bsn} and BMN\cite{exper:bmn}.

In the benchmark evaluation experiment, we consider four classic TAP algorithms (i.e., R-C3D, TURN, BSN and BMN) for the DFES task and then compare their performance on our SDFE-LV database.
In addition, to evaluate the best performance of these algorithms on our database, different types of feature extraction backbone networks are added to the benchmark experiments. Note that, all feature extraction backbone networks are pretrained on the well-performing DFEW database.

Table \ref{benchmark} shows the benchmark experimental results on the SDFE-LV database.
From the upper part of Table \ref{benchmark}, we can clearly see that the bottom-up fashion algorithms (i.e., BSN and BMN) outperform the top-down fashion algorithms (i.e., R-C3D and TURN) by a large margin in AR@AN and AUC.
We speculate that this is due to different proposal generation methods, where top-down models cannot adjust the boundary flexibly because of the distribution limitation of the predefined proposals, while bottom-up models can evaluate the video key expression frames locally and can flexibly combine proposals at the frame level.
The lower part of Table \ref{benchmark} shows the performance comparison of the well-performing BMN under different feature extraction backbones.
Experimental results show that when VGG19-LSTM is used as the feature extraction network, the AUC reaches the highest result of 59.56, and most AR@ANs also achieve the best performance.
Nevertheless, the performance of the DFES task on our database still at a low level, which also reflects the challenge of our database on the DFES task.

\subsection{Duration Challenge}

\begin{figure}[htbp]
\centering
\includegraphics[width=0.95\linewidth]{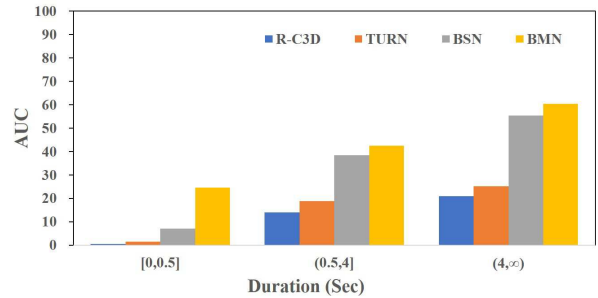}
\caption{Comparison results of R-C3D, TURN, BSN and BMN on expression clips with different durations.}
\label{fig:exp1}
\end{figure}

In order to further analyze the difficulty of the expression spotting task in SDFE-LV, we explore the spotting ability of four models for complete expression clips with different durations.
Figure \ref{fig:exp1} shows the spotting ability of the four algorithms for expression clips with different durations. The experimental results illustrate that BSN and BMN algorithms outperform R-C3D and TURN, in the spotting expression clips with all different durations by a large margin. 
In addition, we find that it is difficult for all models to spot expression clips with durations less than 0.5 second.
We believe that it is due to the large frame interval, which is the length of the minimum image sequence used for temporal feature extraction. 
However, a smaller frame interval greatly increases the computational cost. 
Therefore we set the frame interval to 16 for all algorithms after weighing the computational cost and model performance. 
In this case, we believe that spotting expression clips of various durations while ensuring high operating efficiency of the algorithm will be another extremely challenging problem of our database.

\subsection{Evaluation on Pretrained Databases and Models}

It is shown in Table \ref{benchmark} that the pretraining of the feature extraction model can greatly affect the performance of DFES task. Therefore, in this subsection, abundant experiments are conducted to explore the influence of pretrained databases and models on the DFES.

\subsubsection{Pretrained Databases}

\begin{table}
\centering
\tiny
\caption{Evaluation on three Pretrained Databases: CK+, AFEW, and DFEW}
\label{databases}
\resizebox{0.9\linewidth}{!}{
\begin{tabular}{lcccc} 
\hline
\hline 
\textbf{Networks} & \textbf{CK+} & \textbf{AFEW} & \textbf{DFEW} \\ 
\hline 
C3D               & 37.03        & 37.98         & \textbf{49.44}         \\
R3D               & 34.34        & 35.35         & \textbf{45.63}         \\
R2Plus1D          & 33.45        & 33.79         & \textbf{46.89}         \\ 
\hline 
Average		  & 34.94		 & 35.71		 & \textbf{47.32}		 \\
\hline
\hline 
\end{tabular}}
\end{table}

Table \ref{databases} shows the spotting AUC of BMN on SDFE-LV with the feature encoding model pretrained by the CK+, AFEW and DFEW databases, where CK+ is a lab-controlled database containing 327 labeled expression sequences, AFEW and DFEW are in-the-wild databases with 1,156 and 11,697 emotional videos.
We find that using the AFEW pretrained C3D model can improve the performance of expression spotting to a certain extent compared to CK+ with similar sample magnitudes.
In addition, we also find that when pretrained with in-the-wild DFEW with larger sample size, the performance of C3D achieves a greater improvement, which is 12.41 and 11.46 higher than CK+ and AFEW, respectively.
Experimental results of the R3D\cite{exper:r3d} and R2Plus1D\cite{exper:R2Plus1D} feature encoding models on three FER databases also reveal the same conclusion.
Therefore, we can conclude that pretrained by large-scale and in-the-wild FER databases can effectively improve the spotting performance of various models on the SDFE-LV database.

%

\subsubsection{Pretrained Models}

\begin{table}
\centering
\caption{Evaluation on Pretrained Models}
\label{models}
\resizebox{\linewidth}{!}{
\begin{tabular}{lcccc} 
\hline
\hline 
\textbf{Networks} & \textbf{CK+} & \textbf{AFEW} & \textbf{DFEW}  & \textbf{Average} \\ 
\midrule
R3D               & 34.34        & 35.35         & 45.63           & 38.44             \\
Resnet18-LSTM     & 46.89        & 47.90         & 56.43           & 50.41             \\
Resnet50-LSTM     & 47.47        & 46.68         & 59.03           & 51.06             \\
Resnet101-LSTM    & 41.54        & 48.21         & 59.19           & 49.65             \\
\midrule
C3D               & 37.03        & 37.98         & 49.44           & 41.48             \\
VGG13-LSTM        & 50.59        & 51.11         & 58.79           & 53.50             \\
VGG16-LSTM        & 50.63        & 52.05         & 59.47           & 54.05             \\
VGG19-LSTM        & 52.70        & 53.05         & 59.56           &55.10		       \\ 
\midrule
Two TSN (Resnet18)  & 48.18        & 48.97         & 56.24           & 51.13             \\ 
Two TSN (Resnet50)  & 49.53        & 48.44         & 57.15           & 51.71             \\ 
Two TSN (Resnet101) & 47.89        & 46.09         & 57.06           & 50.35             \\ 
\hline
\hline 
\end{tabular}}
\end{table}

Table \ref{models} shows the spotting AUC of BMN on SDFE-LV with different constructions of pretrained feature extraction models.
Comparing Resnet-LSTM networks and VGG-LSTM networks with R3D and C3D respectively, we find that the spotting performance of the 2D-LSTM networks far outperform the 3D networks.
In addition, we also explore the performance of TSN\cite{exper:tsn}, a widely used feature extraction network in the field of action, which operates on single RGB frame and stacked optical flow field to capture appearance feature and motion information.
Experimental results show that TSN still has great advantages compared to 3D networks, which is 12.5 higher on average.
We believe this is because the stepwise and separate modeling of spatial and temporal by 2D-LSTM networks and TSN can better perceive facial expression movements in long videos than the simultaneous modeling of 3D models in spatial and temporal.

\subsection{Context Information Evaluation}

\begin{table}
\centering
\caption{Context Information Evaluation}  
\label{context}
\resizebox{\linewidth}{!}{
\begin{tabular}{lcccc} 
\hline
\hline 
\textbf{Networks} & \textbf{CK+} & \textbf{AFEW} & \textbf{DFEW}  & \textbf{Average} \\ 
\midrule
C3D               & 37.03        & 37.98         & 49.44           & \textbf{41.48}             \\
Two C3D           & 36.03        & 37.72         & 50.69           & \textbf{41.48}             \\
\midrule
Resnet50-LSTM     & 47.47        & 46.68         & 59.03           & \textbf{51.06}             \\
Two Resnet50-LSTM & 45.38        & 47.21         & 59.15           & 50.58             \\
\midrule
VGG19-LSTM        & 52.70        & 53.05         & 59.56           & \textbf{55.10}    \\ 
Two VGG19-LSTM    & 47.68        & 51.61         & 57.76           & 52.35             \\
\hline
\hline 
\end{tabular}}
\end{table}
Several studies have proven that FER performance can be significantly improved by considering context information, such as gestures and places \cite{exper:context2, exper:context1, intro:caer}.
Therefore, we compare the performance of single and two stream networks (i.e., C3D, Resnet50-LSTM and VGG19-LSTM) on SDFE-LV in Table \ref{context} by AUC.
Surprisingly, experimental results illustrate that the addition of context stream not only does not improve the performance of the model in expression spotting, but also significantly reduces the performance.
We speculate this is because the viewing angles in our SDFE-LV videos are relatively fixed or slowly changing, which makes it difficult to establish temporal correspondence between background and subtle expression changes, and excessive attention to the context stream will degrade the spotting performance on SDFE-LV. 


\section{Conclusion and Discussion}

In this paper, a large-scale, multi-source, and unconstrained long video-based dynamic facial expression database called SDFE-LV has been presented. SDFE-LV aims to facilitate the development of dynamic facial expression spotting (DFES), which is a challenging research topic in the fields of affective computing and artificial intelligence, while less attention has been paid in communities. It is worthy to mention that the long videos in SDFE-LV were collected from multiple real-world/closely real-world media sources, e.g., interviews, documentaries, movies, and we-media short videos, which facilitates the dynamic facial expression analysis close to requirement in real life applications. Extensive baseline experiments were conducted on this database. Experimental results revealed that DFES issue is indeed challenging and need to be further explored by more researchers from the view of artificial intelligence and affective computing. In the future, we will continue to expand the database by collecting more samples from real life environments, and explore advanced spotting algorithms to tackle the DFES task.

\bibliography{SDFE-LV}

\end{document}